\documentclass[letterpaper, 10 pt, conference]{ieeeconf}  

\IEEEoverridecommandlockouts                              

\usepackage{graphicx} 
\usepackage{siunitx}  
\usepackage{cite}
\usepackage[caption=false]{subfig}
\usepackage{booktabs}
\usepackage{xfrac}
\usepackage{amsmath,amssymb,amsfonts}
\usepackage{array,multirow,graphicx}
\usepackage[hyphens]{url}
\usepackage{hyperref}
\usepackage{tikz}

\newcommand{\mycomment}[1]{}
\newcommand\copyrighttext{
	\footnotesize 
	\textcopyright~2023 IEEE. Personal use of this material is permitted. Permission from IEEE must be obtained for all other uses, in any current or future media, including reprinting/republishing this material for advertising or promotional purposes, creating new collective works, for resale or redistribution to servers or lists, or reuse of any copyrighted component of this work in other works. DOI to be provided shortly.%
	}%
\newcommand\copyrightnotice{%
    \begin{tikzpicture}[remember picture,overlay]%
 	\node[anchor=south, xshift=0pt, yshift=4pt] at (current page.south)%
 	{\fbox{\parbox{\dimexpr\textwidth-\fboxsep-\fboxrule\relax}{\copyrighttext}}};%
 	\end{tikzpicture}%
}%

\title{\LARGE \bf
Advancing Frame-Dropping in Multi-Object Tracking-by-Detection Systems Through Event-Based Detection Triggering
}

\author{Matti Henning, Michael Buchholz, and Klaus Dietmayer
\thanks{M. Henning,  M. Buchholz, and K. Dietmayer are with the Institute of Measurement, Control, and Microtechnology of the University of Ulm, 89081, Ulm, Germany. E-Mail: {\tt\small \{firstname.lastname\}@uni-ulm.de}.}%
\thanks{This research was accomplished within the projects UNICAR\emph{agil} (FKZ
16EMO0290) and AUTOtech.\emph{agil} (FKZ 01IS22088W).
We acknowledge the financial support for the projects by the
German Federal Ministry of Education and Research (BMBF).}
}

\begin{document}

\maketitle
\thispagestyle{empty}
\pagestyle{empty}

\begin{abstract}
With rising computational requirements modern automated vehicles (AVs) often consider trade-offs between energy consumption and perception performance, potentially jeopardizing their safe operation. 
Frame-dropping in tracking-by-detection perception systems presents a promising approach, although late traffic participant detection might be induced.

In this paper, we extend our previous work on frame-dropping in tracking-by-detection perception systems. We introduce an additional event-based triggering mechanism using camera object detections to increase both the system's efficiency, as well as its safety.  
Evaluating both single and multi-modal tracking methods we show that late object detections are mitigated while the potential for reduced energy consumption is significantly increased, reaching nearly 60 Watt per reduced point in HOTA score. 
\end{abstract}

\section{INTRODUCTION}
\copyrightnotice%
The safe operation of automated vehicles (AVs) is both a complex, as well as a thoroughly researched topic. 
One of the central elements for safe operation is the vehicle's environment perception which enables the AV to properly react to changing environment conditions. 
Consequently, modern AVs often employ multi-modal and multi-redundant sensor setups~\cite{Buchholz2021} to achieve an accurate representation of the environment.
The generated sensor data of this setup is commonly processed by deep-learning-based object detection models running on high-end GPUs.
In this way, the environment perception task induces significant computational and resource requirements that reduce the operation time of AVs. 
Further, the complexity and the high processing time of the employed object detection models can lead to significant reaction delay of the AV, ultimately jeopardizing its safe operation~\cite{becker2020, betz2023, toschi2019}. 
\begin{figure}[!t]
    \centering
    \includegraphics[width=.9\linewidth]{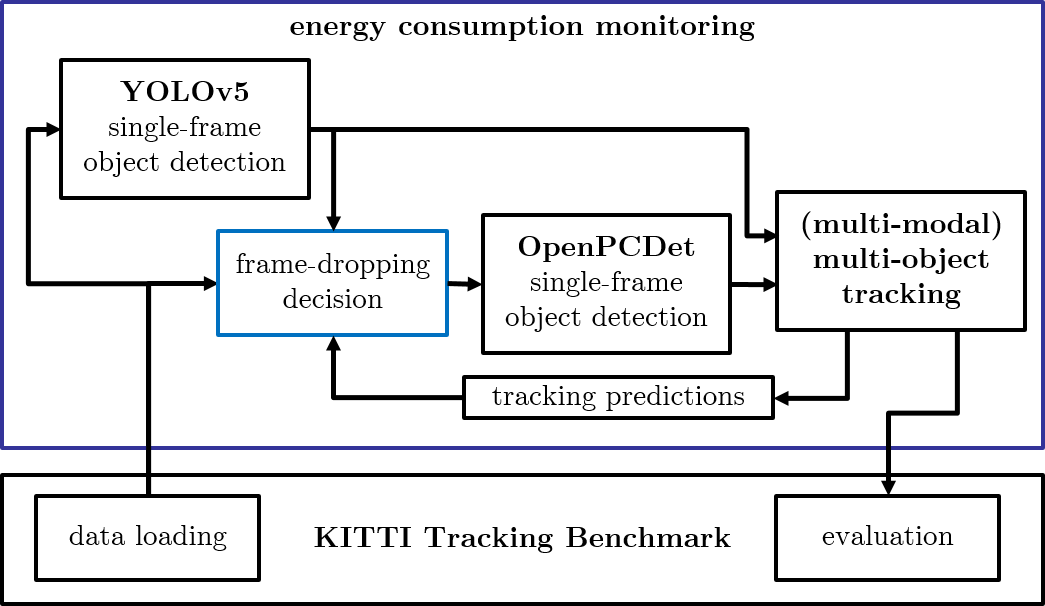}
    \caption{Interaction between dataset, object detection, frame-dropping decision, and tracking. Three lidar object detection models are evaluated independently (cf. Section~\ref{sec:method}).}
    \label{fig:intro:overview}
\end{figure}


Various approaches to alleviate these effects exist, aiming to either reduce the complexity or increase the efficiency of the environment perception. 
These approaches both focus on individual elements within the environment perception system, e.g., by employing architectural changes to neural networks~\cite{liang2021}, as well as on the consolidated environment perception system, e.g., by considering the current situation of the vehicle~\cite{henning2022saep}. Here, the AV's situation is leveraged to reconfigure the environment perception system in an optimal manner, achieving significant reductions in energy consumption. 
Within this context, we have recently presented an approach for scalable employment of deep-learning-based lidar object detection models in tracking-by-detection systems using frame-dropping~\cite{henning2023framedrop}. 
In this work, major reductions in energy consumption are achieved at minor performance loss. 
However, the achieved performance significantly degrades at object detection frame rates below \SI[mode=text]{5}{\hertz}. 
Further, newly appearing objects can be detected late at low frame rates, potentially leading to risk-inducing behavior of the AV. 

In this work, we extend our previously presented approach in~\cite{henning2023framedrop} by leveraging the low inference time and high accuracy of camera-based object detection.
Specifically, we evaluate the discrepancy between the predicted object states from the previous frame with the object camera detections of the current frame. If the discrepancy is significant, the data processing of the current frame for lidar object detection is triggered in addition to the rigid frame-dropping parameterization employed in~\cite{henning2023framedrop}, effectively presenting an event-based triggering of data processing.
In this manner, the concerns of a delayed reaction to changing environment conditions, e.g., a late object detection, are alleviated, while also extending the range of application of our previously presented approach. 
An overview of our method is presented in Fig.~\ref{fig:intro:overview}, which is further detailed in Section~\ref{sec:method}.

Our main contributions can be summarized as follows:
\begin{itemize}
    \item We extend our previously presented approach for frame-dropping in tracking-by-detection perception systems by an additional event-based triggering method for lidar-based object detection and integrate both camera object detection, as well as multi-modal tracking.
    \item We show that the resulting perception performance and efficiency are significantly increased at low frame rates and that the risk due to late object detection is mitigated. 
\end{itemize}

\section{Related Work}
As we extend our previous work from~\cite{henning2023framedrop} this section briefly summarizes the approach. Further, our extension relates to event-based triggering, which we will briefly put into context.

In ~\cite{henning2023framedrop} the disparity of computational requirements between object detection and object tracking in tracking-by-detection systems is leveraged.
Specifically, the model-based stabilization capability of the tracking step is used to compensate for dropped frames in the detection step. 
By dropping the data processing in the detection step the overall energy consumption of the perception system can be significantly reduced. In this manner, scalable employment of usually rigid deep-learning-based 3D lidar object detection models is enabled without any requirement for model adaptations.
The work evaluates a selection of object detection models using OpenPCDet~\cite{openpcdet2020} and the single-sensor 3D multi-object tracking framework CasTrack~\cite{CasTrack, Wu2021}. 
The evaluation is conducted on the KITTI Tracking Benchmark dataset\cite{Geiger2012CVPR} using fixed frame-dropping rates ranging from processing every frame to only processing one out of ten frames. It is shown that a significant reduction in the perception systems energy consumption can be achieved at a reasonable decline in perception performance, reaching a yield of up to \SI[mode=text]{15.0}{}\si{\watt} per reduced point in HOTA~\cite{Luiten2020} score at a processing target of \SI[mode=text]{50}{\percent}, i.e., processing every second frame. However, reducing the processing target further results in a sharp decline in achieved perception performance, especially in detection accuracy. This is also shown in a qualitative case of late object detection for an object appearing out of occlusion. 

To alleviate these effects, in this work, we introduce a method to enable additional out-of-sequence triggering of the lidar object detection step. By evaluating the discrepancy between predicted objects and detected camera objects (cf. Section~\ref{sec:method}) we effectively introduce an event-based processing trigger, i.e., in the event of a significant discrepancy we trigger the additional processing of the current frame.
This event-based triggering approach has found application in distributed sensor networks to reduce energy-intensive communication overhead~\cite{trimpe2014}. Further,~\cite{noack2022} leverages event-based triggering in a tracking application to increase communication efficiency and resilience.
However, using an event-based approach to trigger data processing of another sensor modality for multi-modal object tracking applications is, to the best of our knowledge, unprecedented.

\section{Method}
\label{sec:method}
The conceptual foundation of this work is presented in~\cite{henning2023framedrop}. 
In this Section, the elements of our method extension, as presented in Fig.~\ref{fig:intro:overview}, are further detailed, and its contributions are highlighted against our previous approach.

\subsection{Employed Frameworks}
\label{sec:method:frameworks}
The core of our extended approach remains a tracking-by-detection system. 
Consequently, we continue to employ OpenPCDet~\cite{openpcdet2020} providing single-frame 3D lidar object detections using the models PV-RCNN~\cite{shi2020_pvrcnn}, SECOND~\cite{yan2018second}, and PointPillars~\cite{lang2019pointpillars}.
The previously used Point-RCNN is neglected due to its comparably low performance at high energy consumption.
Further, to generate single-frame camera object detections we employ YOLOv5\cite{jocher2022, schoen2023}, providing excellent object detection performance and inference speed.

The generated object detections are fed to the multi-object tracking element of our approach. Here, we continue to employ CasTrack~\cite{CasTrack} as a single-sensor tracking framework. 
For CasTrack the generated camera object detections are used solely for the purpose of frame-dropping decision-making.
To enable the full potential of the generated camera object detections, i.e., using the generated detections also for the state estimation of the employed tracking framework, we additionally integrate DeepFusionMOT~\cite{Wang2022} as an open-source multi-modal tracking framework. In this manner, both lidar and camera object detections influence the performance of the evaluated tracking-by-detection system. 

For the evaluation of our extended method (cf. Section~\ref{sec:eval}) the combinations of lidar object detection models and tracking variants are considered. Both tracking variants are adapted to take frame-dropping (cf. Section~\ref{sec:method:drop}) into account.
Verification of the core functionality as in~\cite{henning2023framedrop} using available ground-truth label data, representing perfect object detection methods, is omitted for brevity. 
 
\subsection{Data Handling}
\label{sec:method:data}
\begin{table}[t]
    \centering
    \caption{Applied baseline target to process $n$ out of $m$ frames.}
    \begin{tabular}{rcccccc}
    \toprule
    processing target & \SI[mode=text]{100}{\percent} & \SI[mode=text]{50}{\percent} & \SI[mode=text]{33}{\percent} & \SI[mode=text]{20}{\percent} & \SI[mode=text]{10}{\percent}\\
    \sfrac{$n$}{$m$}  & \sfrac{1}{1}  & \sfrac{1}{2} &  \sfrac{1}{3} & \sfrac{1}{5} & \sfrac{1}{10}\\
    \bottomrule
    \end{tabular}
    \label{tab:method:subsampling}
\end{table}
The presented perception systems (cf. Section~\ref{sec:method:frameworks}) are used to process data of the KITTI Tracking Benchmark dataset~\cite{Geiger2012CVPR}. The dataset consists of 21 training sequences and 29 test sequences. Due to the lack of label availability for the test set, we evaluate our method on the validation split of the training sequences. 
The data frames are sequentially fed to the evaluated perception system and respective processing results are stored for post-processing evaluation of the perception performance (cf. Section~\ref{sec:method:eval}). 
Further, to ensure a fair comparison between the evaluated perception system's energy consumption, we adhere to the emulated cycle time of  $t=\SI[mode=text]{100}{\milli\second}$ between frames introduced in~\cite{henning2023framedrop}. In this manner, a dropped data frame in the lidar object detection step will correctly reduce the energy consumption during the duration of sequence data processing.

As per~\cite{henning2023framedrop} we employ pre-defined continuous frame-dropping for the lidar detection step as a baseline processing target.
Building on the results presented in~\cite{henning2023framedrop} we focus on lower frame rates and define the applied frame-dropping baselines as per Table~\ref{tab:method:subsampling}. 
The baseline processing targets refer to the continuous frame-dropping decision that can be overwritten by the discrepancy evaluation (cf. Section~\ref{sec:method:drop}). Similarly to~\cite{henning2023framedrop}, the applied frame-dropping is referred to by its processing target for the remainder of this work.

\subsection{Frame-Dropping Decision}
\label{sec:method:drop}
\begin{figure}[!t]
    \centering
    \includegraphics[width=\linewidth]{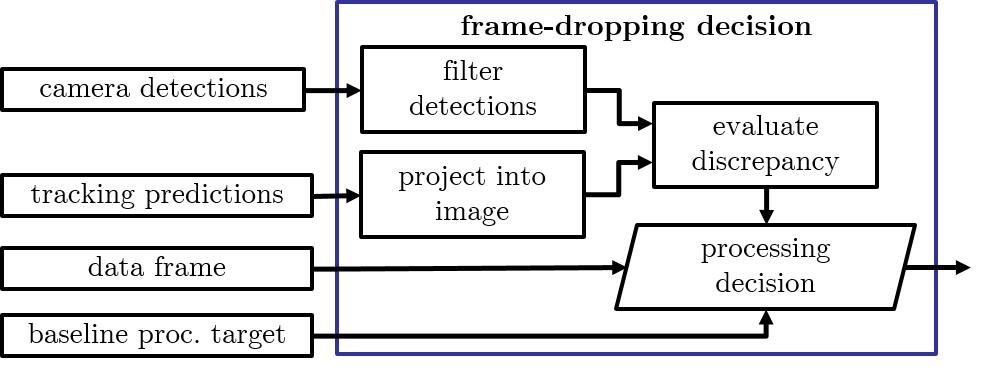}
    \caption{Method overview to determine the frame-dropping decision.}
    \label{fig:method:drop}
\end{figure}
Next to the continuous frame-dropping decision as per~\cite{henning2023framedrop} (cf. Section~\ref{sec:method:data}) we introduce an additional discrepancy evaluation that can overwrite the continuous decision to drop a frame for lidar object detection processing. This effectively enforces the processing of a data frame even though it would normally have been dropped as per the continuous processing target.
The method of this case is outlined in Fig.~\ref{fig:method:drop}.

For the current data frame, the camera object detections are generated. As we specifically focus on improving the safety of our approach w.r.t. changing environments in close proximity, we employ a simple projection filter for the generated camera object detections. 
We use the camera calibration data to estimate the approximate distance $d_i$ of a camera object detection $i$
\begin{align}
    d_i = \frac{\text{approximate height} \cdot \text {focal length}}{\text{pixel height}_i}
\end{align}
according to a class-specific approximate object height in \si{\meter} and the detected object pixel height. In this manner, by filtering object detections over a parameterizable distance threshold $d_{\text{max}}$, the high sensitivity of the camera object detector is alleviated, so that enforced processing of data frames due to camera object detections outside of the direct proximity of the AV is avoided.

In parallel, the predicted object states from the previous frame, generated by the employed tracking framework, are projected into the camera image as 2D bounding boxes. 
Using the filtered camera object detections and the projected 2D lidar object detections the 2D Box Intersection-over-Union (IoU) is evaluated to determine the forwarding decision. 
Here, if for any of the filtered camera objects no sufficient match can be found in the predicted objects, i.e., any IoU lies below $\text{IoU}_\text{min}$, processing of the lidar data frame is enforced.

The result of this approach is an increased number of processed frames compared to the applied baseline processing target. This is referred to as the effective processing target and indicated accordingly in Section~\ref{sec:eval}. 

\subsection{Evaluation}
\label{sec:method:eval}
With the method extension of this work, we aim to improve the perception performance. We especially focus on low frame rates, i.e., low baseline processing targets (cf. Section~\ref{sec:method:data}), aiming to mitigate the outlined shortcomings of the previous approach. For that matter, we continue to leverage the tight integration of the evaluation methods~\cite{luiten2020trackeval} provided by CasTrack and present the evaluation results based on HOTA~\cite{Luiten2020} and CLEAR~\cite{bernardin2008} metrics.

Further, we evaluate the overall system power draw using an external measurement device (cf.~\cite{henning2023framedrop}), to identify the processing requirements of our extended approach. We continue to use the \textit{yield}~\cite{henning2023framedrop},
\begin{align}
    \text{yield}^\text{target}_\text{model} &= \frac{\text{system draw}_\text{model}^\text{100}-\text{system draw}_\text{model}^\text{target}}{\text{HOTA}_\text{model}^\text{100}-\text{HOTA}_\text{model}^\text{target}}\,,
\end{align}
of an evaluated perception system.
The yield enables a comparison between perception systems, representing the reduction in the system power draw in \si{\watt} per reduced point in HOTA score in relation to the respective baseline at \SI[mode=text]{100}{\percent} baseline processing target.

The introduced metrics HOTA, CLEAR, and yield refer to better behaviors at larger values. Only the system draw reflects better behavior at smaller values. 

\subsection{Limitations}
\label{sec:method:limits}
Our extension relies on the employed camera object detection model. Here, three main limitations can be identified.

First, the introduced event-based trigger assumes the correctness of the camera object detections, so that our approach is recall bound by its performance. 
Consequently, it is essential to maintain the baseline processing target as a lower bound to mitigate potential shortcomings in the camera object detector. 
Further, low performance of the camera object detector, e.g., ghost object detections due to sensor degradation, might lead to a high number of processed frames. As this would negate the benefits of frame-dropping we advise sensor degradation monitoring for online applications. 
As no sensor degradations are contained in the evaluated dataset this is out of the scope of this work.

Second, the association between camera objects and tracked objects relies on overlapping sensing areas and assumes observability in both modalities. Occasional partial occlusions can be mitigated by maintaining the baseline processing target. Additionally, multi-camera association adaptations, e.g., as per~\cite{kim2021}, might be employed to compensate for the camera sensors' smaller field of view. Using a single-camera setup in this work, such approaches are omitted. 

Third, the structure of our extended approach (cf. Fig.~\ref{fig:intro:overview} and Fig.~\ref{fig:method:drop}) induces a delay in the perception system by processing the camera detections before processing the lidar data. Using the fast YOLOv5 model lowers the induced delay, especially compared to the inference time of the chosen lidar detection models. 
Further, additional energy is required for processing the camera data. While this is expected for the multi-modal approach of DeepFusionMOT it is counter-intuitive for CasTrack. Within the context of applied and safe automated driving, we assume camera data processing is conducted irrespective of the employed tracking method, e.g., for traffic sign recognition, and object detections are available as a byproduct.
This assumption is backed by~\cite{yeong21}, supporting the necessity of multi-modal perception approaches to avoid the perceptual limitations of the individual sensor modalities.
\begin{table*}[t]
    \centering
    \caption{Evaluation results for the perception systems. Effective processing target in \si{\percent}. MOTA in \si{\percent}, MOTP in \si{\percent}, median system power draw in \si{\watt}, yield in \si{\watt} per reduced point in HOTA score. Highest achieved yield \textbf{bold}, second-highest yield \underline{underlined}.}
    \label{tab:eval:res}
    \setlength\tabcolsep{.1cm}
\begin{tabular}{crlccccccccccccccc}
\toprule
& &&  \multicolumn{5}{c}{PointPillars}&  \multicolumn{5}{c}{PV-RCNN}&  \multicolumn{5}{c}{SECOND}\\
\midrule
\multirow{6}{*}{CasTrack baseline} & \multicolumn{2}{c}{effective proc. target}& 100& 50& 33& 20& 10& 100& 50& 33& 20& 10& 100& 50& 33& 20& 10\\
&$\uparrow$&MOTA& 81.2& 76.2& 53.0& 40.0& 30.9& 83.7& 77.7& 56.6& 42.2& 32.0& 83.0& 76.7& 54.1& 37.2& 29.0\\
&$\uparrow$&MOTP& 87.8& 86.5& 85.5& 83.7& 81.4& 88.6& 87.6& 86.5& 84.7& 82.2& 87.4& 86.5& 85.4& 83.7& 81.0\\
&$\uparrow$&HOTA& 74.9& 70.2& 61.4& 53.1& 42.8& 78.0& 72.9& 63.2& 56.7& 46.0& 77.0& 72.0& 62.6& 55.4& 44.5\\
&$\downarrow$&Sys. draw& 396& 375& 313& 256& 221& 461& 384& 335& 295& 256& 494& 418& 349& 297& 241\\
&$\uparrow$&yield& -& 4.4& 6.2& 6.4& 5.5& -& 15.0& 8.5& 7.8& 6.4& -& 15.3& 10.1& 9.1& 7.8\\
\midrule
\multirow{6}{*}{CasTrack extension} & \multicolumn{2}{c}{effective proc. target}& -& 55& 41& 30& 22& -& 55& 40& 30& 23& -& 54& 39& 29& 21\\
&$\uparrow$&MOTA& -& 76.7& 56.9& 51.6& 45.5& -& 77.9& 60.8& 52.7& 46.7& -& 77.4& 58.6& 50.9& 41.7\\
&$\uparrow$&MOTP& -& 86.6& 85.6& 84.6& 83.7& -& 87.6& 86.6& 85.4& 84.4& -& 86.5& 85.5& 84.3& 83.1\\
&$\uparrow$&HOTA& -& 70.4& 62.5& 57.7& 52.0& -& 73.0& 64.2& 59.7& 54.5& -& 72.2& 63.9& 59.3& 52.3\\
&$\downarrow$&Sys. draw& -& 378& 319& 266& 230& -& 388& 346& 306& 270& -& 423& 354& 302& 260\\
&$\uparrow$&yield& -& 7.3& 4.1& 6.2& 7.6& -& 8.1& 14.3& 8.3& 8.5& -& 9.5& 14.9& 10.7& 10.8\\
\midrule
& &&  \multicolumn{5}{c}{PointPillars}&  \multicolumn{5}{c}{PV-RCNN}&  \multicolumn{5}{c}{SECOND}\\
\midrule
\multirow{6}{*}{DeepFusionMOT baseline} & \multicolumn{2}{c}{effective proc. target}& 100& 50& 33& 20& 10& 100& 50& 33& 20& 10& 100& 50& 33& 20& 10\\
&$\uparrow$&MOTA& 76.5& 75.6& 64.7& 44.7& 5.4& 74.2& 71.7& 62.2& 43.4& 4.5& 70.8& 67.4& 58.1& 36.4& 0.0\\
&$\uparrow$&MOTP& 78.2& 77.8& 77.1& 76.2& 76.1& 78.9& 78.5& 77.8& 76.6& 76.2& 78.2& 77.7& 77.0& 75.9& 75.5\\
&$\uparrow$&HOTA& 66.0& 64.6& 58.9& 50.9& 37.4& 66.5& 65.0& 59.5& 52.1& 39.4& 64.6& 63.2& 58.4& 50.7& 37.9\\
&$\downarrow$&Sys. draw& 399& 381& 315& 259& 225& 464& 385& 338& 295& 261& 477& 417& 349& 297& 247\\
&$\uparrow$&yield& -& 13.1& 11.8& 9.3& 6.1& -& \underline{52.7}& 18.0& 11.8& 7.5& -& 43.5& 20.4& 12.9& 8.6\\
\midrule
\multirow{6}{*}{DeepFusionMOT extension} & \multicolumn{2}{c}{effective proc. target}& -& 54& 39& 27& 19& -& 55& 39& 28& 20& -& 53& 38& 27& 18\\
&$\uparrow$&MOTA& -& 76.0& 69.4& 64.2& 50.5& -& 72.6& 66.4& 61.0& 47.1& -& 68.8& 63.3& 55.7& 40.5\\
&$\uparrow$&MOTP& -& 77.8& 77.4& 76.9& 76.0& -& 78.5& 78.0& 77.4& 76.7& -& 77.7& 77.2& 76.6& 75.6\\
&$\uparrow$&HOTA& -& 64.8& 60.9& 58.1& 51.4& -& 65.2& 61.1& 58.9& 52.7& -& 63.6& 60.0& 57.0& 51.2\\
&$\downarrow$&Sys. draw& -& 379& 316& 264& 228& -& 389& 342& 300& 266& -& 421& 355& 305& 261\\
&$\uparrow$&yield& -& 11.7& 16.9& 16.5& 17.1& -& 14.4& \textbf{59.2}& 22.8& 21.6& -& 16.1& \underline{52.7}& 26.5& 22.4\\
\bottomrule
\end{tabular}
\end{table*}
\section{Evaluation}
\label{sec:eval}

This section provides the evaluation results of our extended method and compares them to the baseline method from~\cite{henning2023framedrop}. The results are consolidated in Table~\ref{tab:eval:res}. Analogous to~\cite{henning2023framedrop}, to generate the evaluation results the dataset is processed in Python on a consumer-grade PC running Ubuntu. The system is equipped with an AMD Ryzen Threadripper 2990WX CPU and an Nvidia RTX 2080Ti 11GB GPU on 64GB RAM.

The parameterization of our method extension is empirically chosen. The approximate height is set to \SI[mode=text]{1.5}{\meter}, $d_\text{max}$ is set to \SI[mode=text]{25}{\meter}, and $\text{IoU}_\text{min}$ is set to 0.25.

\subsection{Performance of the Perception Framework}
\begin{figure}[t!]
    \centering
    \subfloat[HOTA score vs. system power draw using CasTrack (CT).]{
    \includegraphics[width=.97\linewidth]{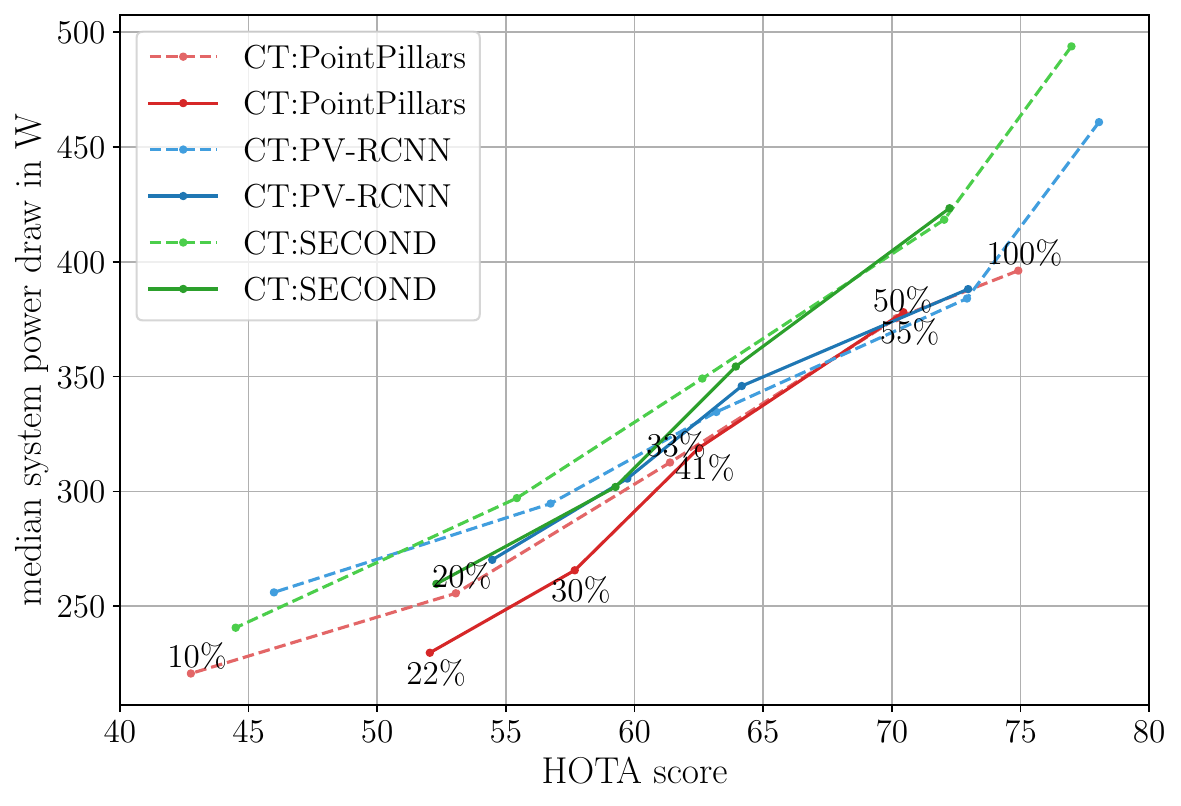}
    \label{fig:eval:hota:CT}}
    \hfil
    \subfloat[HOTA score vs. system power draw using DeepFusionMOT (DF).]{
        \includegraphics[width=.97\linewidth]{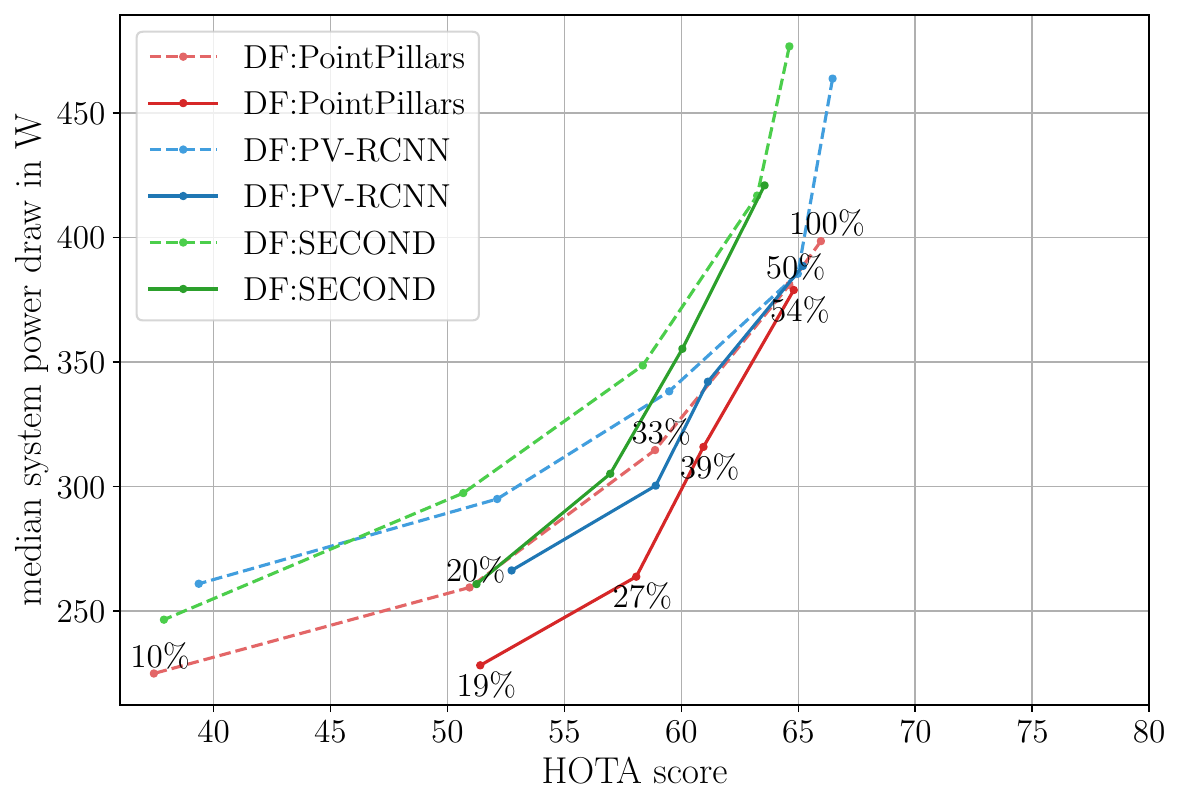}
        \label{fig:eval:hota:DF}}
    \hfil
    \caption{Achieved HOTA score vs. induced system draw for the detection model variants from Table II. Dashed lines refer to the baseline frame-dropping variant. Solid lines refer to the extension of this work. Percentages refer to effective processing targets. For readability, only the processing targets of PointPillars are indicated.}
    \label{fig:eval:hota}
\end{figure}

The results presented in Table~\ref{tab:eval:res} are visually consolidated in Fig.~\ref{fig:eval:hota} regarding the perception performance at different processing targets.
The figures~\ref{fig:eval:hota:CT} and~\ref{fig:eval:hota:DF} present the results for the respective tracking variants CasTrack and DeepFusionMOT.
Interestingly, the multi-modal perception systems using DeepFusionMOT are outperformed by the single-sensor perception systems using CasTrack. Although this is initially counter-intuitive,~\cite{pang2020} confirms that multi-modal approaches often achieve lower performance compared to single-modality approaches. 
They present a late-fusion approach to mitigate these shortcomings, which is out of the scope of this work.
While the highest achieved performance results for the perception systems using DeepFusionMOT are similar between the three evaluated detection models, PV-RCNN achieves the highest results in a perception system using CasTrack. For both tracking variants SECOND is the most expensive, followed by PV-RCNN. PointPillars is the least energy-intensive variant, although it achieves respectable perception performance.

Comparing the presented extension results to the baseline results, Fig.~\ref{fig:eval:hota} shows that our extension increases the perception system's efficiency in nearly all cases of baseline processing targets below \SI[mode=text]{50}{\percent}. Conferring Table~\ref{tab:eval:res}, the performance increase corresponding to our method extension largely relates to an increase in the MOTA value. 
This confirms the functionality of our event-based triggering approach for environmental changes in close proximity to the AV.
Overall, the presented extension results, indicated by solid lines in figures~\ref{fig:eval:hota:CT} and~\ref{fig:eval:hota:DF}, achieve a higher HOTA score at their respective median system draw compared to simply increasing the baseline processing target, indicated as dashed lines. 
As our presented method increases the number of processed frames it is expected that the system draw increases as well. 
The presented results confirm that our method extension benefits both the performance, as well as the energy consumption of the presented perception systems at lower processing targets. 
Consequently, the achieved results of our method extension (solid lines) are further to the lower-right of the figures, reflecting a higher HOTA score at a lower median system draw.
Only PV-RCNN at \SI[mode=text]{33}{\percent} baseline processing target using CasTrack achieves slightly better efficiency without our extension. This likely corresponds to the overall good performance of the detection model and the induced overhead of camera object detection for the single-modality tracking approach.
Further, for baseline processing targets of \SI[mode=text]{50}{\percent} the achieved HOTA score and resulting system draw are practically identical, supporting our focus on lower baseline processing targets.

For the two employed tracking variants, Fig.~\ref{fig:eval:hota} shows that the increase in efficiency is considerably larger for perception systems employing DeepFusionMOT. This is according to our expectation, as the camera object detections are used in subsequently dropped frames to stabilize newly generated objects from additionally processed frames as per our method extension. 
Perception systems employing CasTrack still benefit from our extension, verifying the effectiveness of model-based prediction capabilities at low frame rates.

\begin{figure}[!t]
    \centering
    \includegraphics[width=.96\linewidth]{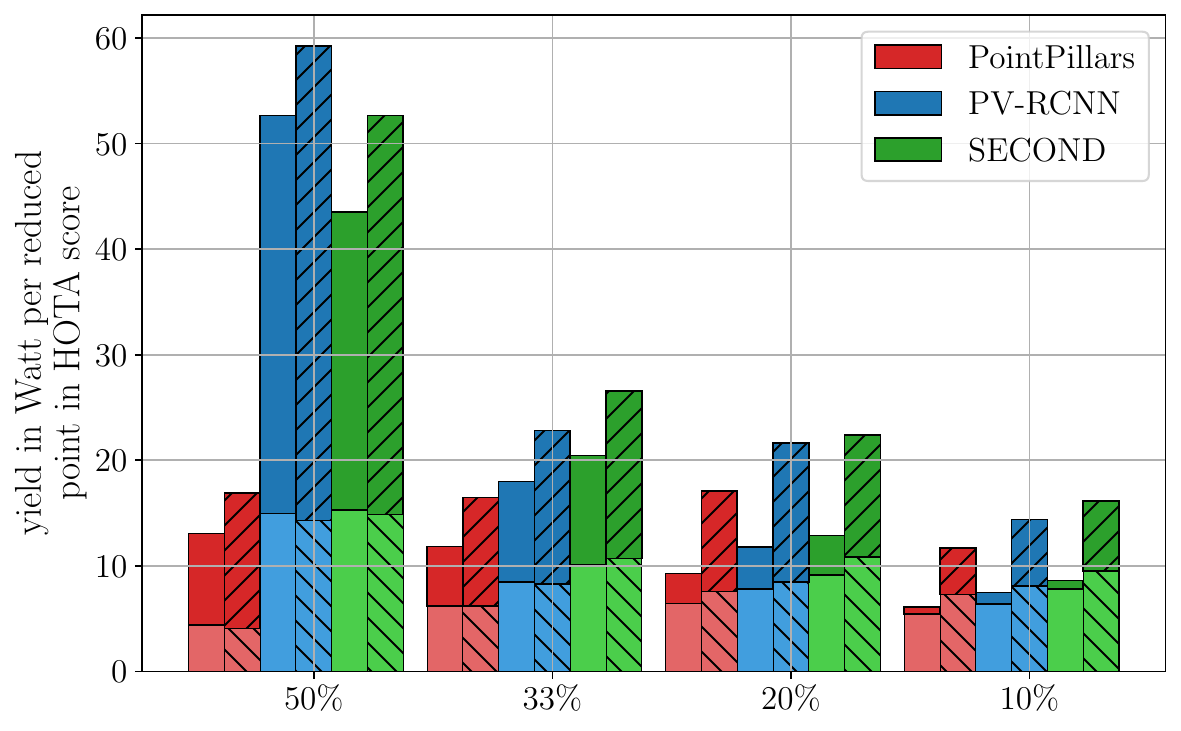}
    \caption{Achieved yield for the evaluated perception systems at the indicated baseline processing targets. Systems using CasTrack and systems using DeepFusionMOT are indicated by reduced and full saturation respectively. Hatched elements refer to the frame-dropping extension of this work.}
    \label{fig:eval:yield}
\end{figure}

Supporting the results from Fig.~\ref{fig:eval:hota} the yield of the evaluated perception systems, i.e., the reduction in system draw per reduced point in HOTA score, is presented in Fig.~\ref{fig:eval:yield}. The results are grouped as per baseline processing target, although the effective processing targets are increased by the extension of this work (cf. Table~\ref{tab:eval:res}).

The results for perception systems using CasTrack (reduced saturation) verify that the introduced extension (hatched elements) improves the perception systems efficiency at lower baseline processing targets, but introduces a slight overhead at \SI[mode=text]{50}{\percent}, as well as at \SI[mode=text]{33}{\percent} for PV-RCNN. 
Further, the resulting yield remains roughly similar throughout the evaluated processing targets, which matches our expectation that a single-modality setup is ill-suited to maximize the efficiency of our multi-modal approach.
For perception systems using DeepFusionMOT (full saturation), the achieved yield of the baseline processing target as per~\cite{henning2023framedrop} already outperforms the achieved yield for perception systems using CasTrack. The improvement declines with the baseline processing target reflecting the lowered detection accuracy.
The results of the event-based triggering extension presented in this work (hatched elements) show significant improvements in the achieved yield throughout all evaluated baseline processing targets, reaching a yield of up to \SI[mode=text]{59.2}{\watt} per reduced point in HOTA score. The improvement is also better retained at lower baseline processing targets.
We thereby conclude that our method is well-suited to improve the efficiency of multi-modal tracking-by-detection systems using frame-dropping.

\subsection{Mitigating Late Object Detection}
\begin{figure}[t!]
    \centering
    \subfloat[First frame with a generated object track using the method from~\cite{henning2023framedrop}.]{
    \includegraphics[width=.94\linewidth]{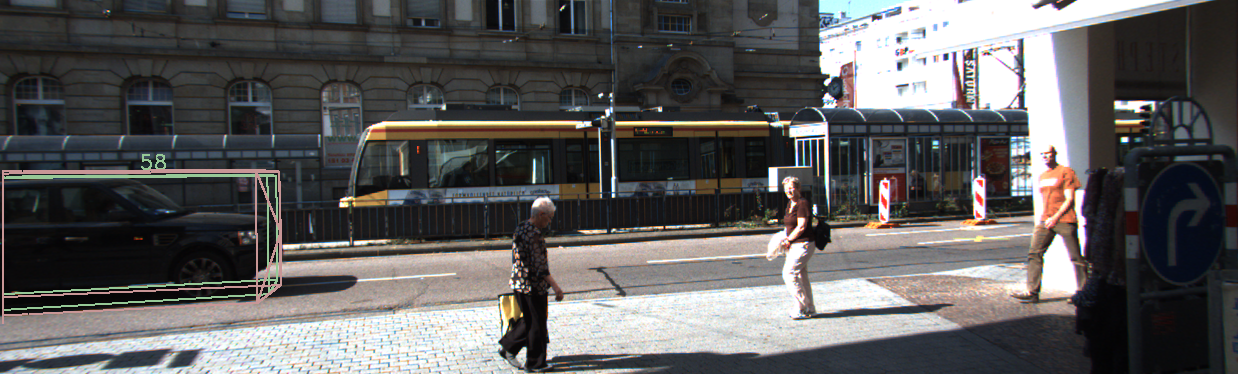}
    \label{fig:eval:latedetection:late}}
    \hfil
    \subfloat[First frame with a generated object track using the presented extension.]{
        \includegraphics[width=.94\linewidth]{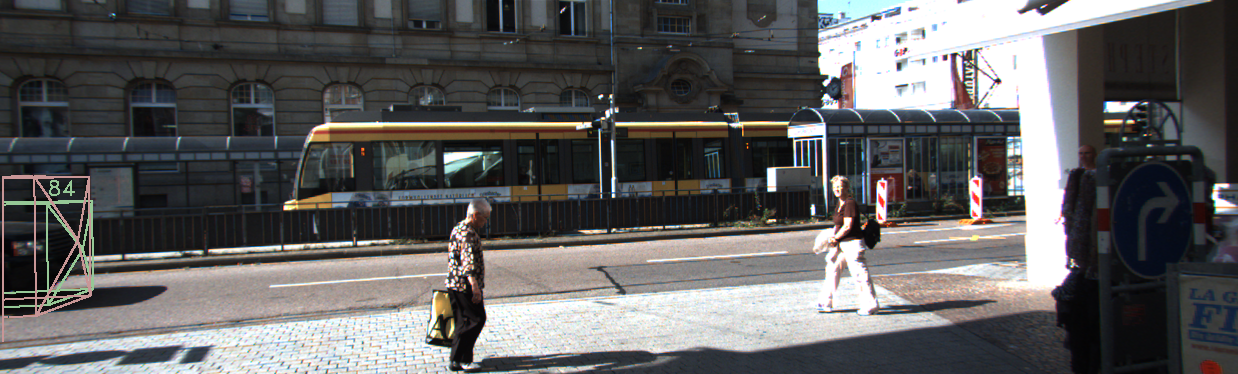}
        \label{fig:eval:latedetection:early}}
    \hfil
    \caption{Example of a potentially risk-inducing situation due to late object detection at \SI[mode=text]{10}{\percent} baseline processing target. Object labels for cars are indicated in red and tracked objects are indicated in green.}
    \label{fig:eval:latedetection}
\end{figure}
Lastly, we reevaluate the potentially risk-inducing scenario due to late object detection from~\cite{henning2023framedrop}.  
A vehicle enters the perception field out of an occlusion on the left-hand side.
The comparison of the baseline method and the extended method of this work are shown in Fig.~\ref{fig:eval:latedetection}.
In Fig.~\ref{fig:eval:latedetection:late} the object is first detected in frame 940. Using the extended method of this work, as shown in Fig.~\ref{fig:eval:latedetection:early}, the object is already detected in frame 936, effectively increasing the reaction time by 4 frames. 
However, our method relies on an evaluation of the camera data before processing the lidar data (cf. Fig~\ref{fig:intro:overview}). 
The induced median delay due to the YOLOv5 inference time is \SI[mode=text]{10.2}{\milli \second}. 
Although this delay is not negligible, it is reasonable both in comparison to the common \SI[mode=text]{100}{\milli \second} real-time assumption and sensor frame rate, as well as in inference time of the respective lidar object detection models. 
As per the presented results on performance and energy consumption in Table~\ref{tab:eval:res}, we conclude that our method extension using an event-based triggering approach significantly contributes to an AVs environment perception efficiency and safety.

\section{Conclusions}
In this work, we have extended our method previously presented in~\cite{henning2023framedrop}, which introduces frame-dropping in multi-object tracking, with an event-based triggering component. Our method extension enforces the additional processing of lidar data in the event of an identified discrepancy between the predicted objects and the detected camera objects. 

Using the foundation of~\cite{henning2023framedrop}, we employed open-source object perception methods for lidar and camera data and extended the framework with an additional multi-modal multi-object tracking framework. 
Evaluating our method extension in comparison to the baseline using the KITTI Tracking Benchmark dataset we have shown that significant performance improvements can be achieved at baseline processing targets below \SI[mode=text]{50}{\percent} while increasing the corresponding system power draw only marginally. Overall, we increase the yield from up to \SI[mode=text]{15.3}{\watt} per reduced point in HOTA score as per the baseline in~\cite{henning2023framedrop} to up to \SI[mode=text]{59.2}{\watt} using our presented event-based trigger extension. 
Further, we showed that the potential safety risk of late object detection can be fully mitigated. This enables the usage of lower baseline processing targets to fully leverage the potential to reduce the energy consumption of the perception system.

With this work, we have established a strong baseline between detection methods, as well as single and multi-modality tracking-by-detection systems. In our continued work, we aim to improve the performance of multi-modal perception systems and to apply our approach to online closed-loop automated driving in urban applications.
\bibliographystyle{IEEEtran}%
\bibliography{IEEEabrv,root}%
\end{document}